\documentclass[conference,a4paper]{APSIPA2025}
\usepackage{amsmath}
\usepackage{graphicx}
\usepackage{multirow}
\usepackage{threeparttable}
\usepackage[numbers,sort&compress]{natbib}

\usepackage{amssymb}
\usepackage{algorithm}
\usepackage{algpseudocode}
\usepackage{mathtools}
\usepackage{booktabs}
\usepackage{wrapfig}
\usepackage[table]{xcolor}
\definecolor{mygreen}{HTML}{3CB371}
\definecolor{myblue}{HTML}{6495ED}
\definecolor{myred}{HTML}{CD5C5C}
\definecolor{mypurple}{HTML}{9370DB}
\definecolor{whitebox}{HTML}{D9D9D9}

\usepackage{geometry}
\usepackage{fancyhdr}

\fancypagestyle{firststyle}{
  \fancyhf{}
  \fancyhead[C]{2025 Asia Pacific Signal and Information Processing Association Annual Summit and Conference (APSIPA ASC)}
}

\fancypagestyle{plain}{
  \fancyhf{}  % clear header/footer

}

\begin{document}

% \title{Uncolorable Examples: Preventing Illegitimate AI Colorization through Adversarial Attacks}
\title{Uncolorable Examples: Preventing Unauthorized AI Colorization
via Perception-Aware Chroma-Restrictive Perturbation}
\author{
\authorblockN{
Yuki Nii\authorrefmark{1},
Futa Waseda\authorrefmark{1}, 
Ching-Chun Chang\authorrefmark{2}, 
Isao Echizen\authorrefmark{1}\authorrefmark{2}
}

% $^\ddag$

\authorblockA{
\authorrefmark{1}
The University of Tokyo, Japan \\
E-mail: \{yuki-nii, futa-waseda\}@g.ecc.u-tokyo.ac.jp}

\authorblockA{
\authorrefmark{2}
National Institute of Informatics, Japan \\
E-mail: \{ccchang, iechizen\}@nii.ac.jp}
}

\maketitle
\thispagestyle{firststyle}

\pagestyle{plain}

\begin{abstract}
% AI colorization is a powerful and visually impressive technology.
% However, it poses risks of copyright infringement—e.g., the unauthorized colorization and resale of monochrome manga and films.
% Despite these concerns, no effective method currently exists to prevent such misuse, highlighting the need for protective mechanisms.
% We introduce \textit{Uncolorable Examples}, the first defense designed to prevent unauthorized image colorization. To ensure real-world applicability, we define four key criteria for such defense: effectiveness, imperceptibility, transferability, and robustness.
% Our method, \textbf{Perception-Aware Chroma-Restrictive Perturbation (PAChroma)}, generates Uncolorable Examples by crafting adversarial perturbations guided by a Laplacian filter and strengthened through data augmentation.
% Experiments across multiple datasets demonstrate that PAChroma significantly degrades colorization quality while maintaining the visual appearance.
% This work represents the first step toward protecting visual content from illegitimate AI colorization, paving the way for copyright-aware defenses in generative media.

% Modified by Waseda
% 165 words
AI-based colorization has shown remarkable capability in generating realistic color images from grayscale inputs.
However, it poses risks of copyright infringement—e.g., the unauthorized colorization and resale of monochrome manga and films.
Despite these concerns, no effective method currently exists to prevent such misuse.
To address this, we introduce the first defensive paradigm, \textit{Uncolorable Examples}, which embed imperceptible perturbations into grayscale images to invalidate unauthorized colorization.
To ensure real-world applicability, we establish four criteria: effectiveness, imperceptibility, transferability, and robustness.
Our method, \textbf{Perception-Aware Chroma-Restrictive Perturbation (PAChroma)}, generates Uncolorable Examples that meet these four criteria by optimizing imperceptible perturbations with a Laplacian filter to preserve perceptual quality, and applying diverse input transformations during optimization to enhance transferability across models and robustness against common post-processing (e.g., compression). 
Experiments on ImageNet and Danbooru datasets demonstrate that PAChroma effectively degrades colorization quality while maintaining the visual appearance.
This work marks the first step toward protecting visual content from illegitimate AI colorization, paving the way for copyright-aware defenses in generative media.

\end{abstract}

\section{Introduction}

% (Follow abstract)
% (Motivation:)
% Colored images offer greater visual understanding and appeal compared to monochrome images. 
Recent advances in AI colorization~\cite{huang2022unicolorunifiedframeworkmultimodal,wu2022vividdiverseimagecolorization} has demonstrated remarkable capability in generating realistic color images from grayscale inputs.
%(Problem:)
However, these advancements raise significant ethical and legal concerns. In Japan, for instance, a man was arrested for selling unauthorized colorized versions of the famous animation “Godzilla”~\cite{mainichi2025godzilla}. With the increasing accessibility of powerful colorization models, malicious users can easily colorize manga or movies without the creator’s consent and resell them, leading to copyright infringement.
% (Need:)
Yet, no method currently exists to prevent such unauthorized colorization, highlighting the urgent need for protection.

\begin{figure*}
    \centering
    \includegraphics[width=1.0\linewidth]{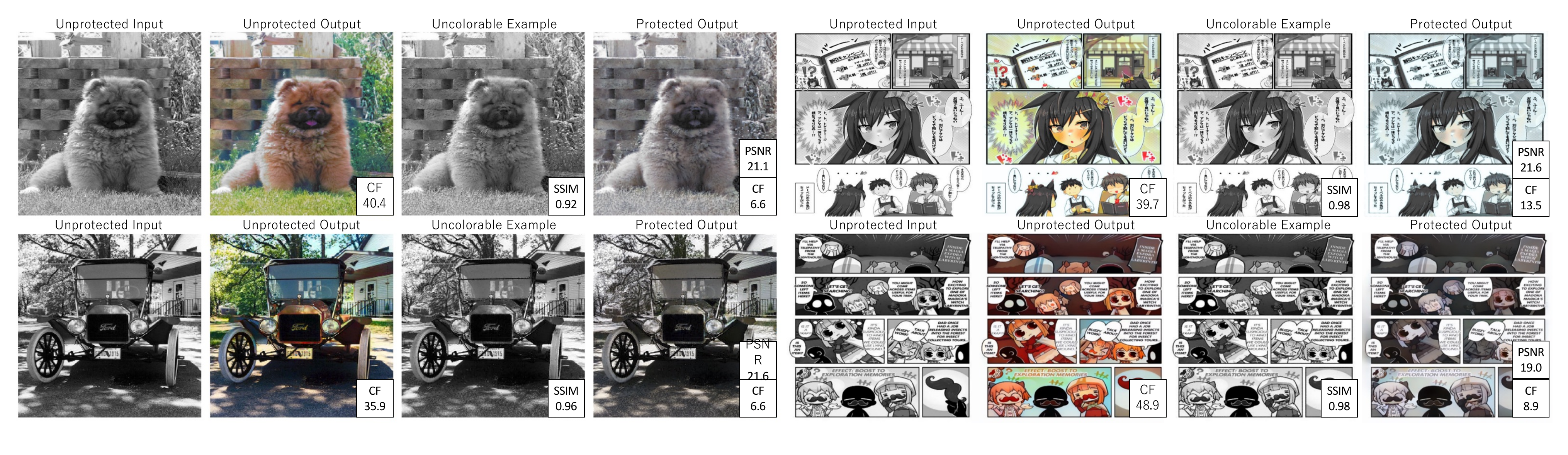}
    \caption{\textbf{Uncolorable Examples from PAChroma invalidate colorization via imperceptible perturbations}. Top-left: DeOldify~\cite{DeOldify2021}; top-right: ACDO~\cite{dakini2024animecolor}; bottom-left: DDColor~\cite{kang2023ddcolorphotorealisticimagecolorization}; bottom-right: MC-V2~\cite{mangacolorv2_2022}. Each image is shown with its CF score, SSIM between the input, or PSNR between the output. }
    \label{fig:example_results}
\end{figure*}

% (Contribution:)
In this paper, we present the first defensive paradigm against unauthorized image colorization, termed \textbf{Uncolorable Examples}, and establish four key criteria for practical applicability: effectiveness, imperceptibility, transferability, and robustness. To meet these criteria, Uncolorable Examples are generated using our proposed method, \textbf{Perception-Aware Chroma-Restrictive Perturbation (PAChroma)}.
PAChroma utilizes the idea of adversarial examples by embedding imperceptible perturbations into grayscale images to invalidate AI colorization (Fig.~\ref{fig:Method}). These perturbations are designed to concentrate on high-frequency regions by leveraging a Laplacian filter to target coarse areas. 
Furthermore, to enhance transferability across diverse colorization models and improve robustness to common image transformations (e.g., resizing, compression), we optimize the perturbations via iterative input transformations.
% (Effectiveness:)
Our method achieves significant suppression of colorization quality with minimal visual change compared to the unprotected image.
% (Experiments:)
We evaluate our method on both natural and manga image datasets, demonstrating consistent effectiveness across multiple state-of-the-art colorization models (Fig.~\ref{fig:example_results}). 
Our contributions are summarized as follows:
\begin{figure}
    \centering
    \includegraphics[width=\linewidth]{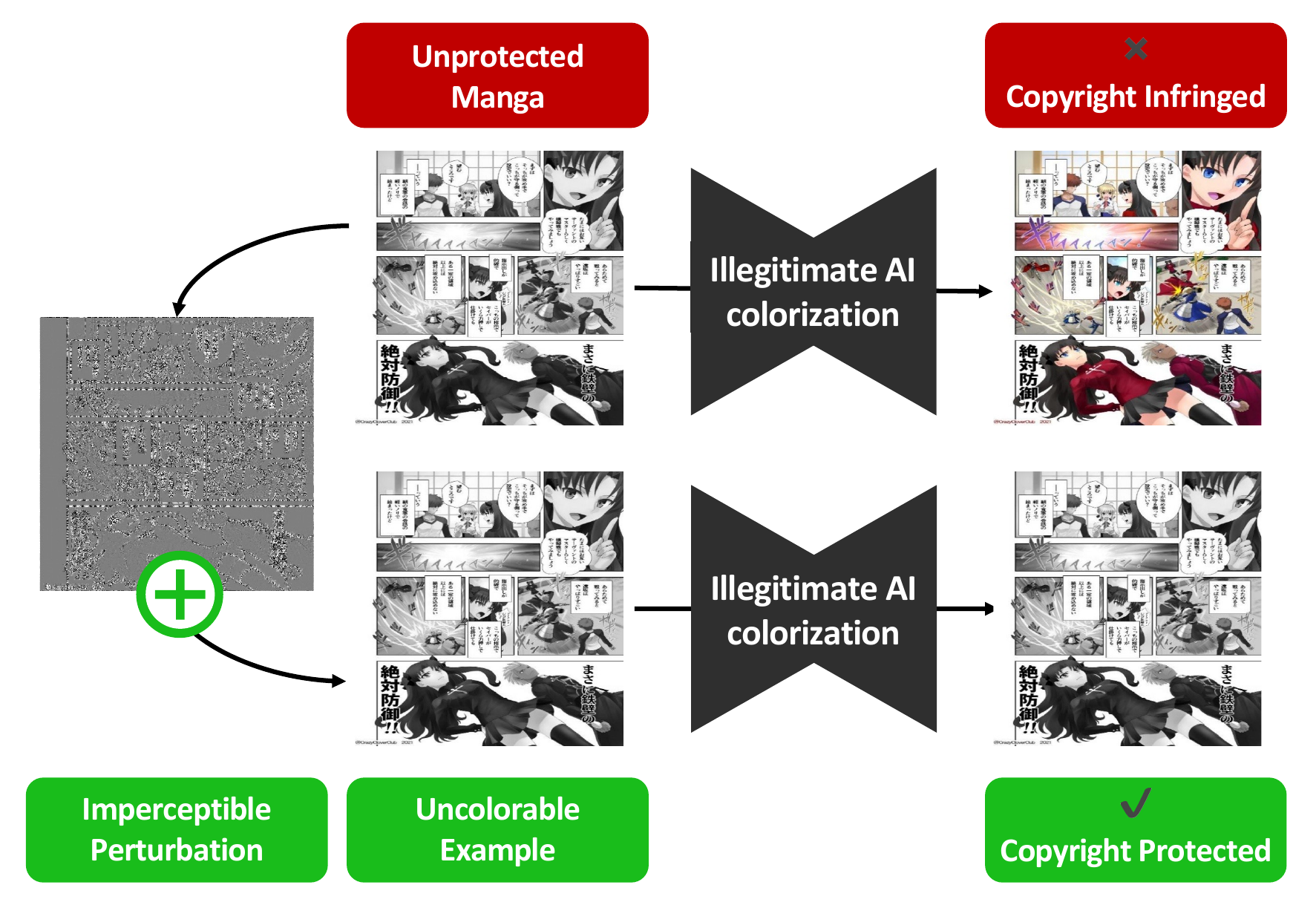}
    \caption{Overview of Uncolorable Examples. Without protection, grayscale images can be illegitimately colorized by AI colorization models. Our method, PAChroma, generates Uncolorable Examples by adding human-imperceptible perturbations to the input, effectively invalidating unauthorized colorization.}
    \label{fig:Method}
\end{figure}

\begin{itemize}
    % \item We are the first to propose \textit{Uncolorable Examples}—grayscale images protected against unauthorized colorization. These are generated using our method, \textit{PAChroma} (Perception-Aware Chroma-Restrictive Perturbation), which leverages adversarial perturbations guided by Laplacian filtering and iterative data augmentation.
    % \item We demonstrate the effectiveness of our approach across state-of-the-art colorization models, achieving imperceptibility, transferability across different model architectures, and robustness to image transformations.
    \item \textbf{Novel defensive paradigm}: We introduce \textit{Uncolorable Examples}, the first defense against unauthorized colorization via imperceptible perturbations to grayscale images.
    \item \textbf{Definition of defense criteria}: We establish four key criteria, \textit{effectiveness}, \textit{imperceptibility}, \textit{transferability}, and \textit{robustness}, for practical colorization defenses.
    \item \textbf{Practical method (PAChroma)}: We propose Perception-Aware Chroma-Restrictive Perturbation (PAChroma), which generates Uncolorable Examples by balancing the four criteria through Laplacian filtering and input transformations during optimization.
    \item \textbf{Comprehensive empirical validation}: Experiments on ImageNet and Danbooru demonstrate that PAChroma effectively invalidates state-of-the-art colorization models without changing the visual quality of the image.
\end{itemize}

\section{Related Work}

\textbf{Automatic Colorization.}
Early work on automatic image colorization focused on CNN-based methods (e.g., DeOldify~\cite{DeOldify2021}) which predict plausible color channels from grayscale inputs. Later, GAN-based methods (e.g., BigColor~\cite{kim2022bigcolorcolorizationusinggenerative} and GCP~\cite{wu2022vividdiverseimagecolorization}) emerged, aiming to produce more diverse and vibrant colorizations using learned priors. More recently, transformer-based approaches (e.g., DDColor~\cite{kang2023ddcolorphotorealisticimagecolorization}) leverage global attention for generating semantically consistent and photorealistic results.
%In this work, we propose a unified attack framework that targets a range of these architectures—including CNN-, GAN-, and transformer-based models.
In this work, we aim at invalidating colorization models—including CNN-, GAN-, and transformer-based models.

\textbf{Adversarial Examples.} 
Adversarial examples are carefully crafted inputs that cause neural networks to produce incorrect results. Szegedy et al.~\cite{szegedy2013intriguing} first revealed the vulnerability of AI models to small perturbations. Among the most well-known attacks, Projected Gradient Descent (PGD)~\cite{madry2018pgd} is regarded as a strong first-order adversary and is widely used. % as a benchmark for adversarial robustness.
While most adversarial examples are used to attack classification models~\cite{dong2018boostingadversarialattacksmomentum},~\cite{goodfellow2015explainingharnessingadversarialexamples}, we repurpose them as a defense mechanism by crafting Uncolorable Examples—grayscale images with imperceptible perturbations that look unchanged to humans, but block AI models from adding unauthorized colors.
% In our study, these adversarial perturbations are optimized to disrupt the colorization process, effectively preserving the original grayscale appearance.
% We use PGD as the simplest baseline for comparison.

\textbf{Image Translation Protection.}
Yeh et al.~\cite{9096939} proposed an attack against GAN-based image translation (e.g., deepfake synthesis) by adding adversarial perturbations to the input image to nullify or distort the model’s result. 
% CMUA~\cite{huang2021cmuawatermarkcrossmodeluniversaladversarial} generates image-universal, cross-model perturbations that can attack multiple models simultaneously, rather than a single target.
% Anti-Forgery~\cite{wang2022antiforgerystealthyrobustdeepfake} crafts adversarial perturbations that are sparse, perceptually natural, and robust to diverse image transformations.
Consequently, subsequent studies~\cite{huang2021cmuawatermarkcrossmodeluniversaladversarial, wang2022antiforgerystealthyrobustdeepfake} have focused on disrupting deepfake generation.
% Inspired by this, we aim to nullify colorization models beyond GANs—specifically targeting CNN- and transformer-based architectures—to ensure transferability across diverse models while keeping the perturbations imperceptible.b
Motivated by this, we aim at preventing unauthorized colorization, which has been unexplored. We introduce a novel defense paradigm, define key criteria, then propose a practical defense.

\section{A Defense Paradigm: Uncolorable Examples}
We introduce a novel defensive paradigm, \textit{Uncolorable Examples}, which embeds imperceptible adversarial perturbations into grayscale images to invalidate unauthorized colorization.
Following~\cite{9096939}, there are two possible strategies for invalidating colorization: (1) nullifying it to produce a grayscale result, and (2) distorting it to produce unnatural colors. The diversity of plausible colorizations makes unnatural outputs unreliable as a defense. We instead steer results toward grayscale to suppress colorization.

Unlike natural images, manga often contains large, flat regions with minimal detail, making perturbations—especially in backgrounds or speech bubbles—visually conspicuous. This highlights the importance of imperceptible defenses. Moreover, colorization models vary widely in architecture, demanding transferable defense, and simple post-processing can easily remove perturbations~\cite{xie2018mitigatingadversarialeffectsrandomization, guo2018counteringadversarialimagesusing}, further emphasizing the need for robustness.
To ensure practical applicability, we propose four criteria an effective defense should satisfy:
\begin{itemize}
    \item \textbf{\textcolor{mygreen}{Effectiveness.}} Perturbations should disable the model's ability to add color, resulting in grayscale outputs.
    \item \textbf{\textcolor{myblue}{Imperceptibility.}} Perturbations should be visually imperceptible to the human eye.
    \item \textbf{\textcolor{mypurple}{Transferability.}} Perturbations should remain effective across different colorization models.
    \item \textbf{\textcolor{myred}{Robustness.}} Perturbations should remain effective under common image transformations (e.g., resizing, cropping and JPEG compression).
\end{itemize}

% We evaluate these four criteria and compare our proposed method, PAChroma, with the Nullifying Attack~\cite{9096939} as a baseline, optimized using the loss defined in Equation~\ref{eq:colorfulness}.

% ---------------- Table ----------------
\begin{table*}[t]
\centering
\caption{Performance of Uncolorable Examples on natural images across colorization models. 
PAChroma effectively suppresses colorization while preserving visual quality, balancing the four criteria. Gray highlighted rows indicate white-box settings.}
\label{tab:attack_eval}
\scriptsize
\renewcommand{\arraystretch}{1.2}
\setlength{\tabcolsep}{3pt}
\begin{tabular}{|
>{\centering\arraybackslash}m{0.9cm}|
>{\centering\arraybackslash}m{0.9cm}|
>{\centering\arraybackslash}m{1.8cm}|
>{\columncolor[HTML]{ECF4EC}}c|
>{\columncolor[HTML]{ECF4EC}}c|
>{\columncolor[HTML]{ECF4EC}}c|
>{\columncolor[HTML]{E8F1FB}}c|
>{\columncolor[HTML]{E8F1FB}}c|
>{\columncolor[HTML]{FDEEEA}}c|
>{\columncolor[HTML]{FDEEEA}}c|
>{\columncolor[HTML]{FDEEEA}}c|}
\hline
\multirow{2}{*}{\shortstack{\textbf{Source} \\ \textbf{Model}}} &
\multirow{2}{*}{\shortstack{\textbf{Attack} \\ \textbf{Model}}} &
\multirow{2}{*}{\shortstack{\textbf{Attack} \\ \textbf{Type}}} &
\multicolumn{3}{c|}{\cellcolor[HTML]{ECF4EC}\textbf{Effectiveness}} & 
\multicolumn{2}{c|}{\cellcolor[HTML]{E8F1FB}\textbf{Imperceptibility}} & 
\multicolumn{3}{c|}{\cellcolor[HTML]{FDEEEA}\textbf{Robustness}} \\
\cline{4-11}
& & & 
\shortstack{\textbf{Unprotected} \\ \textbf{CF}} &
\shortstack{\textbf{Protected} \\ \textbf{CF$\downarrow$}} &
\shortstack{\textbf{PSNR} \\ \textbf{(Output)$\downarrow$}} &
\shortstack{\textbf{PSNR} \\ \textbf{(Input)$\uparrow$}} &
\shortstack{\textbf{SSIM} \\ \textbf{(Input)$\uparrow$}} &
\shortstack{\textbf{JPEG} \\ \textbf{75\% CF$\downarrow$}}& 
\shortstack{\textbf{JPEG} \\ \textbf{50\% CF$\downarrow$}}& 
\shortstack{\textbf{RRC}\\ \textbf{CF}$\downarrow$} \\
\hline
% ---------- DeOldify source ----------
\multirow{10}{*}{DeOldify} 
 & \multirow{4}{*}{DeOldify} 
    & Random            & \cellcolor{whitebox}34.15 & \cellcolor{whitebox}26.99 (-20.90\%) & \cellcolor{whitebox}25.70 & \cellcolor{whitebox}28.96 & \cellcolor{whitebox}0.75 & \cellcolor{whitebox}25.00 & \cellcolor{whitebox}24.34 & \cellcolor{whitebox}24.72 \\
\cline{3-11}
 &  & NA               & \cellcolor{whitebox}34.15 & \cellcolor{whitebox}6.53  (-80.90\%) & \cellcolor{whitebox}23.19 & \cellcolor{whitebox}28.41 & \cellcolor{whitebox}0.74 & \cellcolor{whitebox}7.74  & \cellcolor{whitebox}12.71 & \cellcolor{whitebox}8.62 \\

 &  & NA-Mask~(ours)          & \cellcolor{whitebox}34.15 & \cellcolor{whitebox}6.65  (-80.51\%) & \cellcolor{whitebox}24.88 & \cellcolor{whitebox}43.23 & \cellcolor{whitebox}1.00 & \cellcolor{whitebox}20.65 & \cellcolor{whitebox}24.39 & \cellcolor{whitebox}20.38 \\

 &  & \textbf{PAChroma~(ours)}& \cellcolor{whitebox}34.15 & \cellcolor{whitebox}7.38  (-78.40\%) & \cellcolor{whitebox}24.05 & \cellcolor{whitebox}32.59 & \cellcolor{whitebox}0.95 & \cellcolor{whitebox}10.71 & \cellcolor{whitebox}14.10 & \cellcolor{whitebox}9.06 \\
\cline{2-11}
 & \multirow{3}{*}{BigColor} 
    & NA               & 34.15 & 23.22 (-32.00\%) & 23.14 & 27.10 & 0.71 & 21.09 & 20.14 & 21.05 \\
 &  & NA-Mask~(ours)          & 34.15 & 27.20 (-20.30\%) & 28.34 & 32.54 & 0.95 & 26.16 & 25.36 & 25.64 \\
 &  & \textbf{PAChroma~(ours)}& 34.15 & 24.50 (-28.30\%) & 25.66 & 30.28 & 0.91 & 23.17 & 23.04 & 23.50 \\
\cline{2-11}
 & \multirow{3}{*}{DDColor} 
    & NA               & 34.15 & 25.13 (-26.40\%) & 24.75 & 28.33 & 0.74 & 23.65 & 22.92 & 22.98 \\
 &  & NA-Mask~(ours)          & 34.15 & 32.52 (-4.80\%)  & 36.80 & 40.37 & 0.99 & 30.26 & 28.80 & 30.52 \\
 &  & \textbf{PAChroma~(ours)}& 34.15 & 27.42 (-19.70\%) & 28.19 & 32.60 & 0.95 & 26.70 & 26.47 & 26.82 \\
\hline
% ---------- BigColor source ----------
\multirow{10}{*}{BigColor} 
 & \multirow{3}{*}{DeOldify} 
    & NA               & 29.91 & 21.29 (-28.80\%) & 23.03 & 27.20 & 0.70 & 22.16 & 22.50 & 21.26 \\
 &  & NA-Mask~(ours)          & 29.91 & 28.78 (-3.80\%)  & 31.80 & 34.84 & 0.97 & 26.56 & 25.49 & 28.48 \\
 &  & \textbf{PAChroma~(ours)}& 29.91 & 25.84 (-13.60\%) & 26.74 & 30.79 & 0.92 & 24.58 & 23.89 & 26.01 \\
\cline{2-11}
 & \multirow{4}{*}{BigColor} 
    & Random           & \cellcolor{whitebox}29.91 & \cellcolor{whitebox}27.06 (-9.50\%)  & \cellcolor{whitebox}22.77 & \cellcolor{whitebox}28.96 & \cellcolor{whitebox}0.73 & \cellcolor{whitebox}24.79 & \cellcolor{whitebox}24.02 & \cellcolor{whitebox}23.86 \\
\cline{3-11}
 &  & NA               & \cellcolor{whitebox}29.91 & \cellcolor{whitebox}0.79  (-97.40\%) & \cellcolor{whitebox}22.45 & \cellcolor{whitebox}28.28 & \cellcolor{whitebox}0.72 & \cellcolor{whitebox}1.41 & \cellcolor{whitebox}3.68 & \cellcolor{whitebox}2.28 \\

 &  & NA-Mask~(ours)          & \cellcolor{whitebox}29.91 & \cellcolor{whitebox}1.17 (-96.10\%) & \cellcolor{whitebox}24.02 & \cellcolor{whitebox}37.00 & \cellcolor{whitebox}0.98 & \cellcolor{whitebox}6.74 & \cellcolor{whitebox}11.61 & \cellcolor{whitebox}7.39 \\
 
 &  & \textbf{PAChroma~(ours)}& \cellcolor{whitebox}29.91 & \cellcolor{whitebox}5.15 (-82.80\%) & \cellcolor{whitebox}23.64 & \cellcolor{whitebox}32.48 & \cellcolor{whitebox}0.94 & \cellcolor{whitebox}6.29 & \cellcolor{whitebox}7.83 & \cellcolor{whitebox}6.17 \\
\cline{2-11}
 & \multirow{3}{*}{DDColor} 
    & NA               & 29.91 & 20.07 (-32.90\%) & 23.10 & 27.20 & 0.70 & 20.08 & 21.14 & 20.48 \\
 &  & NA-Mask~(ours)          & 29.91 & 28.58 (-4.50\%)  & 31.29 & 34.39 & 0.97 & 26.49 & 25.45 & 28.53 \\
 &  & \textbf{PAChroma~(ours)}& 29.91 & 24.44 (-18.30\%) & 26.60 & 30.72 & 0.92 & 23.36 & 22.97 & 24.65 \\
\hline
% ---------- DDColor source ----------
\multirow{10}{*}{DDColor} 
 & \multirow{3}{*}{DeOldify} 
    & NA               & 36.84 & 36.99 (-0.40\%) & 22.59 & 28.41 & 0.74 & 28.68 & 26.54 & 31.46 \\
 &  & NA-Mask~(ours)          & 36.84 & 35.97 (-2.40\%) & 37.73 & 43.23 & 1.00 & 31.29 & 29.30 & 37.06 \\
 &  & \textbf{PAChroma~(ours)}& 36.84 & 31.32 (-15.00\%) & 26.18 & 32.59 & 0.95 & 28.31 & 27.55 & 32.11 \\
\cline{2-11}
 & \multirow{3}{*}{BigColor} 
    & NA               & 36.84 & 31.47 (-14.60\%) & 21.35 & 27.10 & 0.71 & 26.43 & 25.96 & 30.77 \\
 &  & NA-Mask~(ours)          & 36.84 & 33.71 (-8.50\%)  & 26.59 & 32.54 & 0.95 & 30.49 & 28.91 & 35.17 \\
 &  & \textbf{PAChroma~(ours)}& 36.84 & 27.71 (-24.80\%) & 24.27 & 30.28 & 0.91 & 26.55 & 25.79 & 29.31 \\
\cline{2-11}
 & \multirow{4}{*}{DDColor} 
    & Random           & \cellcolor{whitebox}36.84 & \cellcolor{whitebox}39.50 (+7.20\%)  & \cellcolor{whitebox}22.42 & \cellcolor{whitebox}28.96 & \cellcolor{whitebox}0.75 & \cellcolor{whitebox}29.68 & \cellcolor{whitebox}27.65 & \cellcolor{whitebox}33.52 \\
\cline{3-11}
 &  & NA               & \cellcolor{whitebox}36.84 & \cellcolor{whitebox}1.43 (-96.10\%) & \cellcolor{whitebox}21.14 & \cellcolor{whitebox}28.33 & \cellcolor{whitebox}0.74 & \cellcolor{whitebox}7.41 & \cellcolor{whitebox}16.49 & \cellcolor{whitebox}12.99 \\

 &  & NA-Mask~(ours)          & \cellcolor{whitebox}36.84 & \cellcolor{whitebox}2.47 (-93.30\%) & \cellcolor{whitebox}22.24 & \cellcolor{whitebox}40.37 & \cellcolor{whitebox}0.99 & \cellcolor{whitebox}20.87 & \cellcolor{whitebox}24.41 & \cellcolor{whitebox}25.22 \\

 &  & \textbf{PAChroma~(ours)}& \cellcolor{whitebox}36.84 & \cellcolor{whitebox}7.60 (-79.40\%) & \cellcolor{whitebox}22.02 & \cellcolor{whitebox}32.60 & \cellcolor{whitebox}0.95 & \cellcolor{whitebox}11.81 & \cellcolor{whitebox}14.72 & \cellcolor{whitebox}12.16 \\
\hline
\end{tabular}

\end{table*}

\section{Method}
\subsection{Overview}
Given a grayscale input image $x_l \in \mathbb{R}^{H \times W}$, we add imperceptible perturbation $\delta \in \mathbb{R}^{H \times W}$ to nullify the colorization generator $G: \mathbb{R}^{H \times W} \to \mathbb{R}^{H \times W \times 3}$.
The perturbation is optimized to minimize the colorfulness score~\cite{colorfulness}, which quantifies visual vividness:
\begin{equation}
\mathcal{L}_{\text{CF}} = \text{Colorfulness}\left( G(x_l + \delta) \right)
\label{eq:colorfulness}
\end{equation} 
To improve imperceptibility, transferability, and robustness, we optimize the perturbation using a Laplacian filter and diverse input transformations (see Algorithm~\ref{alg:pachroma}).

\begin{algorithm}[t]

\caption{Perception-Aware Chroma-Restrictive Perturbation (PAChroma)}
\textbf{Input:} Colorization model $G(\cdot)$; colorfulness loss $L_{\text{CF}}$; Laplacian mask $M$; grayscale image $x_l$; max perturbation $\epsilon$; number of iterations $T$; decay factor $\mu$; block split number $s$; number of transformations $N$ \\
%\hspace{1.2em} max perturbation $\epsilon$; number of iterations $T$; decay factor $\mu$; block split number $s$; number of transformations $N$ \\
\textbf{Output:} Final adversarial image $x^{\text{adv}}_T$
\begin{algorithmic}[1]
\State Initialize: $\alpha = \epsilon / 10$, $g_0 = 0$, $x^{\text{adv}}_0 = x_l$
\For{$t = 0$ to $T-1$}
    \State Generate a set of transformed inputs $\mathcal{X} = \{ x^{\text{tran}}_i \}_{i=1}^N$ using structure-invariant transformations
    \For{$i = 1$ to $N$}
        \State Compute colorized output: $x^{\text{rgb}}_i = G(x^{\text{tran}}_i)$
        \State Compute gradient: $g^{(i)} = \nabla_{x} L_{\text{CF}}(x^{\text{rgb}}_i)$
    \EndFor
    \State Compute averaged gradient: $\bar{g}_{t+1} = \frac{1}{N} \sum_{i=1}^N g^{(i)}$
    \State Update momentum: $g_{t+1} = \mu g_t + \frac{\bar{g}_{t+1}}{\| \bar{g}_{t+1} \|_1}$
    \State Compute perturbation step: $\Delta = M \cdot \alpha \cdot \text{sign}(g_{t+1})$
    \State Update adversarial image: $x^{\text{adv}}_{t+1} = \text{Clip}(x^{\text{adv}}_t + \Delta)$

\EndFor
\State \Return $x^{\text{adv}}_T$
\end{algorithmic}
\label{alg:pachroma}
\end{algorithm}

\subsection{Perception-Aware Chroma-Restrictive Perturbation}
We propose Perception-Aware Chroma-Restrictive Perturbation (PAChroma), a method that produces Uncolorable Examples. 
PAChroma applies the Momentum Iterative Fast Gradient Sign Method (MI-FGSM)~\cite{dong2018boostingadversarialattacksmomentum} as its core optimization loop, incorporating the input transformation strategy of Structural Invariant Attack (SIA)~\cite{shen2021structure} and a continuous Laplacian mask during each iteration (Algorithm~\ref{alg:pachroma}).

\subsubsection{Input Transformation}
To enhance transferability and robustness, PAChroma applies structure-preserving augmentations~\cite{shen2021structure}. The input is divided into blocks (e.g., $3\times3$), and random transformations are applied independently to each. 
The transformations include geometric changes (shift, flip, rotation), intensity modifications (scaling, jitter, noise), frequency-domain filtering (DCT), resolution changes (resizing), and spatial dropout ($p{=}0.1$).
At each step, $N$ transformed inputs are used to compute a loss that encourages generalization.

% \begin{figure}
%    \centering
%    \includegraphics[width=1\linewidth]{Images/SIA.jpg}
%     \caption{SIA block-wise transformation}
%     \label{fig:sia}
% \end{figure}

\begin{figure}
    \centering
    \includegraphics[width=1\linewidth]{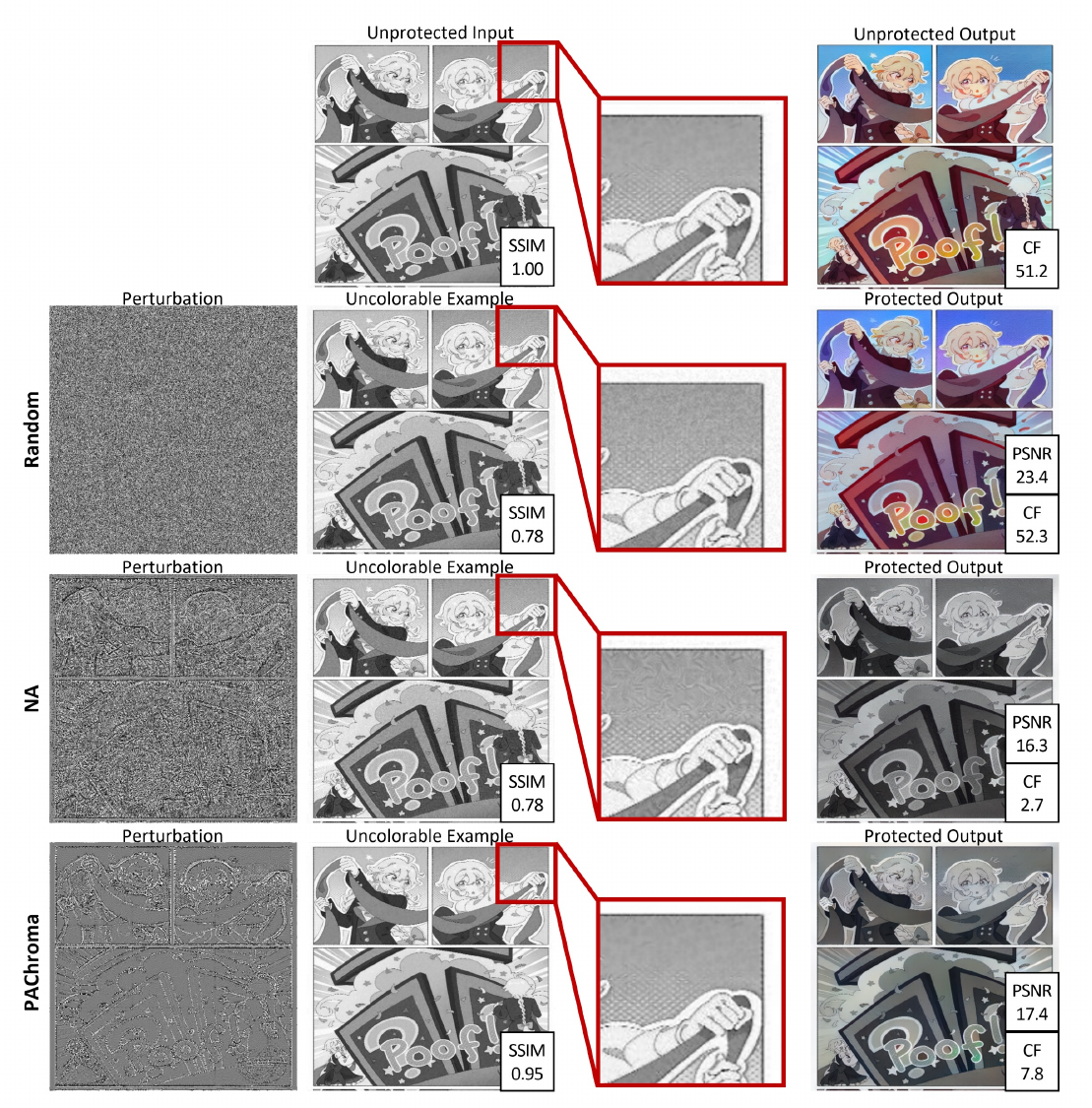}
    \caption{\textbf{Effectiveness and Imperceptibility of random noise, Nullifying Attack (NA), and PAChroma}. PAChroma preserves grayscale structure while preventing colorization, outperforming NA in imperceptibility. Each image includes CF score, SSIM between the input, or PSNR between the output.}
    \label{fig:effect_imper}
\end{figure}

\begin{figure*}[t]
    \centering
    \includegraphics[width=1\linewidth]{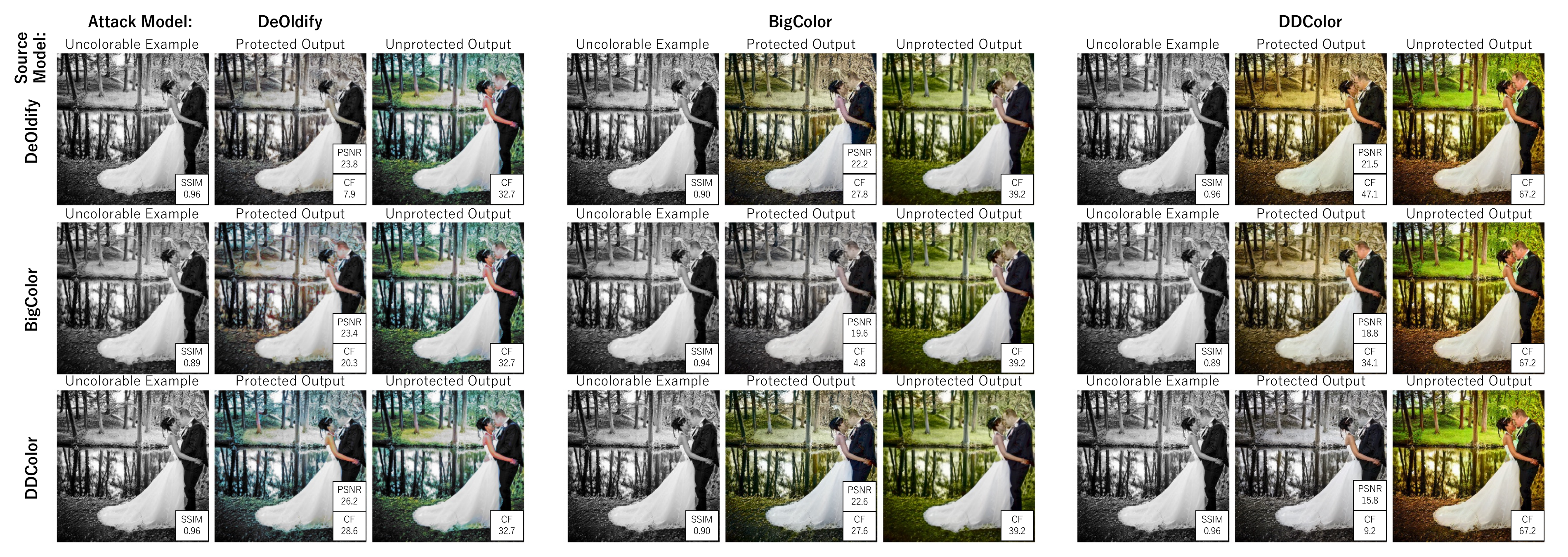}
    
    \caption{\textbf{Transferability of PAChroma} among DeOldify, BigColor, and DDColor. Shown with CF, SSIM between inputs, or PSNR between outputs.}
    \label{fig:trans}
\end{figure*}

\begin{figure}
    \centering
    \includegraphics[width=1\linewidth]{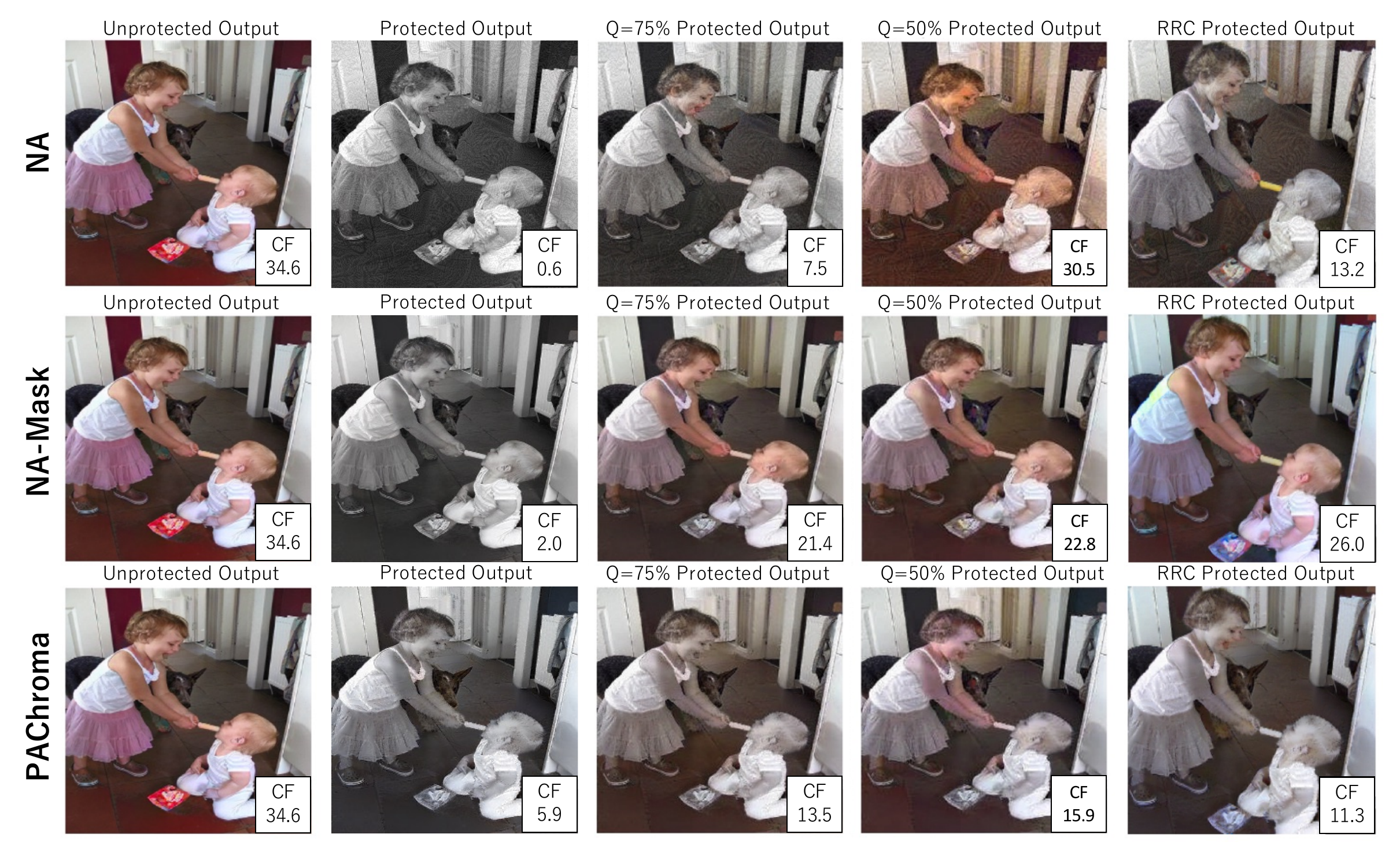}
    \caption{\textbf{Robustness} of NA, NA-Mask, and PAChroma to JPEG compression (Q=$X$\%) and Random Resized Cropping (RRC). Each image is shown with its CF score on the bottom corner.}
    \label{fig:robust}
\end{figure}

\subsubsection{Continuous Laplacian Mask}
To enhance imperceptibility, we guide gradient updates with a continuous Laplacian mask, exploiting reduced distortion visibility along edges due to contrast masking in human vision~\cite{watson1997contrast}.
The continuous Laplacian map $M$ of the input $x_l$ is computed via convolution with a standard Laplacian kernel:
\begin{equation}
M = |\nabla^2 x_l| = \left| x_l * K_{\text{Laplacian}} \right|,
\end{equation}
where $K_{\text{Laplacian}}$ is defined as $[0, 1, 0; 1, -4, 1; 0, 1, 0]$, and $*$ denotes convolution.
The resulting map is normalized to range $[0, 1]$ and applied as a multiplicative weighting mask on the gradient during each update step.

By integrating input transformations with a Laplacian mask, PAChroma produces Uncolorable Examples that are simultaneously effective, imperceptible, transferable, and robust. Visualization results of PAChroma are presented in Fig.~\ref{fig:example_results}.

\section{Experiments}
\subsection{Experimental Setup}
\subsubsection{Colorization Models}
We evaluate our method on three natural image colorization models—CNN-based \textbf{DeOldify}~\cite{DeOldify2021}, GAN-based \textbf{BigColor}~\cite{kim2022bigcolorcolorizationusinggenerative}, and transformer-based \textbf{DDColor}~\cite{kang2023ddcolorphotorealisticimagecolorization}—using official 12M ImageNet weights. For manga, we use two domain-specific models: \textbf{AnimeColorDeOldify (ACDO)}~\cite{dakini2024animecolor} and \textbf{Manga Colorization V2 (MC-V2)}~\cite{mangacolorv2_2022}.

\subsubsection{Datasets}
\textbf{ImageNet:} We sample 100 images from the ImageNet validation set~\cite{deng2009imagenet}, repeated with five random seeds. All images are resized to $256 \times 256$.
\textbf{Danbooru:} Following~\cite{xie-2018-manga}, we collect 4,367  ``manga''-tagged images from Danbooru, resize them to $576 \times 576$, and randomly sample 30 images with five random seeds.

\subsubsection{Evaluation Protocol}
We evaluate our method using standard metrics in image colorization ~\cite{kang2023ddcolorphotorealisticimagecolorization}, ~\cite{kim2022bigcolorcolorizationusinggenerative}. 
Effectiveness is measured by PSNR and Colorfulness (CF) between unprotected and protected outputs; imperceptibility by PSNR and SSIM between the input. We also evaluate black-box transferability and robustness to post-processing (JPEG 75\%, 50\%, and random resized cropping).

\subsubsection{Defense Settings}
We evaluate PAChroma alongside two baselines: the Nullifying Attack (NA)~\cite{9096939} and NA with Laplacian Mask (NA-Mask). 
%We evaluate \textbf{PAChroma} against the Nullifying Attack (NA)~\cite{9096939}, which serves as our baseline method.
Perturbations are $\ell_\infty$-bounded and are optimized using the loss defined in Eq.~\ref{eq:colorfulness}.
The default hyperparameters are:
$\epsilon = \frac{16}{255}$, $\alpha = \frac{1.6}{255}$, number of iterations $T = 100$, and number of transformed images $N = 20$.
% All experiments are conducted on Tesla V100 GPUs.

% ---------------- Manga Table ----------------
\begin{table*}[htbp]
\centering
\caption{Performance of Uncolorable Examples on manga images across colorization models. PAChroma achieves high effectiveness, imperceptibility, and robustness. Highlighted rows are white-box settings.}
\label{tab:manga_eval}
\scriptsize
\renewcommand{\arraystretch}{1.2}
\setlength{\tabcolsep}{3pt}
\begin{tabular}{|
>{\centering\arraybackslash}m{0.8cm}|
>{\centering\arraybackslash}m{0.8cm}|
>{\centering\arraybackslash}m{1.8cm}|
>{\columncolor[HTML]{ECF4EC}}c|
>{\columncolor[HTML]{ECF4EC}}c|
>{\columncolor[HTML]{ECF4EC}}c|
>{\columncolor[HTML]{E8F1FB}}c|
>{\columncolor[HTML]{E8F1FB}}c|
>{\columncolor[HTML]{FDEEEA}}c|
>{\columncolor[HTML]{FDEEEA}}c|
>{\columncolor[HTML]{FDEEEA}}c|}
\hline
\multirow{2}{*}{\shortstack{\textbf{Source} \\ \textbf{Model}}} &
\multirow{2}{*}{\shortstack{\textbf{Attack} \\ \textbf{Model}}} &
\multirow{2}{*}{\shortstack{\textbf{Attack} \\ \textbf{Type}}} &
\multicolumn{3}{c|}{\cellcolor[HTML]{ECF4EC}\textbf{Effectiveness}} &
\multicolumn{2}{c|}{\cellcolor[HTML]{E8F1FB}\textbf{Imperceptibility}} &
\multicolumn{3}{c|}{\cellcolor[HTML]{FDEEEA}\textbf{Robustness}} \\
\cline{4-11}
& & &
\shortstack{\textbf{Unprotected} \\ \textbf{CF}} &
\shortstack{\textbf{Protected} \\ \textbf{CF$\downarrow$}} &
\shortstack{\textbf{PSNR} \\ \textbf{(Output)$\downarrow$}} &
\shortstack{\textbf{PSNR} \\ \textbf{(Input)$\uparrow$}} &
\shortstack{\textbf{SSIM} \\ \textbf{(Input)$\uparrow$}} &
\shortstack{\textbf{JPEG} \\ \textbf{75\% CF$\downarrow$}}& 
\shortstack{\textbf{JPEG} \\ \textbf{50\% CF$\downarrow$}}& 
\shortstack{\textbf{RRC}\\ \textbf{CF}$\downarrow$} \\
\hline
% ------------------ ACDO Source ------------------
\multirow{7}{*}{ACDO} & \multirow{4}{*}{ACDO}
  & Random & \cellcolor{whitebox}45.25 & \cellcolor{whitebox}59.41 (+31.29\%) & \cellcolor{whitebox}22.79 & \cellcolor{whitebox}29.66 & \cellcolor{whitebox}0.76 & \cellcolor{whitebox}60.47 & \cellcolor{whitebox}57.87 & \cellcolor{whitebox}56.39 \\
\cline{3-11}
 & & NA & \cellcolor{whitebox}45.25 & \cellcolor{whitebox}5.86 (-87.05\%) & \cellcolor{whitebox}21.28 & \cellcolor{whitebox}29.42 & \cellcolor{whitebox}0.80 & \cellcolor{whitebox}9.29 & \cellcolor{whitebox}14.49 & \cellcolor{whitebox}21.88 \\
 & & NA-Mask~(ours) & \cellcolor{whitebox}45.25 & \cellcolor{whitebox}7.72 (-82.95\%) & \cellcolor{whitebox}21.93 & \cellcolor{whitebox}36.64 & \cellcolor{whitebox}0.99 & \cellcolor{whitebox}17.73 & \cellcolor{whitebox}12.74 & \cellcolor{whitebox}26.90 \\
 & & \textbf{PAChroma~(ours)} & \cellcolor{whitebox}45.25 & \cellcolor{whitebox}10.89 (-75.94\%) & \cellcolor{whitebox}21.35 & \cellcolor{whitebox}32.82 & \cellcolor{whitebox}0.98 & \cellcolor{whitebox}11.91 & \cellcolor{whitebox}11.64 & \cellcolor{whitebox}13.42 \\
\cline{2-11}
 & \multirow{3}{*}{MC-V2}
   & NA & 45.25 & 54.06 (+19.48\%) & 24.35 & 29.48 & 0.79 & 54.48 & 51.51 & 49.52 \\
 & & NA-Mask~(ours) & 45.25 & 45.90 (+1.44\%) & 34.24 & 34.48 & 0.99 & 46.21 & 41.94 & 43.00 \\
 & & \textbf{PAChroma~(ours)} & 45.25 & 45.28 (+0.08\%) & 31.40 & 32.52 & 0.97 & 45.53 & 41.46 & 42.32 \\
\hline
% ------------------ MC-V2 Source ------------------
\multirow{7}{*}{MC-V2} & \multirow{3}{*}{ACDO}
   & NA & 54.65 & 41.63 (-23.84\%) & 22.24 & 29.42 & 0.80 & 38.56 & 40.32 & 40.21 \\
 & & NA-Mask~(ours) & 54.65 & 53.22 (-2.63\%) & 35.94 & 36.64 & 0.99 & 53.23 & 50.06 & 53.64 \\
 & & \textbf{PAChroma~(ours)} & 54.65 & 51.93 (-4.99\%) & 29.73 & 32.82 & 0.98 & 51.37 & 48.95 & 52.25 \\
\cline{2-11}
 & \multirow{4}{*}{MC-V2}
   & Random & \cellcolor{whitebox}54.65 & \cellcolor{whitebox}46.46 (-15.00\%) & \cellcolor{whitebox}22.18 & \cellcolor{whitebox}29.65 & \cellcolor{whitebox}0.76 & \cellcolor{whitebox}40.39 & \cellcolor{whitebox}36.21 & \cellcolor{whitebox}39.24 \\
\cline{3-11}
 & & NA & \cellcolor{whitebox}54.65 & \cellcolor{whitebox}2.95 (-94.60\%) & \cellcolor{whitebox}16.41 & \cellcolor{whitebox}29.48 & \cellcolor{whitebox}0.79 & \cellcolor{whitebox}8.16 & \cellcolor{whitebox}16.05 & \cellcolor{whitebox}16.65 \\
 & & NA-Mask~(ours) & \cellcolor{whitebox}54.65 & \cellcolor{whitebox}6.01 (-89.01\%) & \cellcolor{whitebox}17.34 & \cellcolor{whitebox}34.48 & \cellcolor{whitebox}0.99 & \cellcolor{whitebox}28.60 & \cellcolor{whitebox}25.43 & \cellcolor{whitebox}29.94 \\
 & & \textbf{PAChroma~(ours)} & \cellcolor{whitebox}54.65 & \cellcolor{whitebox}15.94 (-70.84\%) & \cellcolor{whitebox}17.71 & \cellcolor{whitebox}32.52 & \cellcolor{whitebox}0.97 & \cellcolor{whitebox}20.93 & \cellcolor{whitebox}20.96 & \cellcolor{whitebox}22.00 \\
\hline
\end{tabular}

\end{table*}

\subsection{Results}
%We provide several visual comparisons of grayscale input, perturbations, Uncolorable Exampless, the corresponding colorized outputs, and ground truth. 
\textbf{Effectiveness.}
PAChroma achieves sufficient reduction of CF, producing results that differ from the unprotected outputs.
As shown in Tab.~\ref{tab:attack_eval}, the baseline Nullifying Attack (NA) achieves CF reduction of 80.90\%–97.40\% and PSNR of 21.14–23.19. While PAChroma shows slightly lower CF reduction (78.40\%–82.80\%) and comparable PSNR (22.02–24.05), it still effectively disables colorization and yields outputs perceptually distinct from the unprotected ones—meeting the goal of protection.
Qualitative comparisons with random noise, NA, and PAChroma are presented in Fig.~\ref{fig:effect_imper}.

\textbf{Imperceptibility.}
PAChroma produces Uncolorable Examples that remain visually close to the unprotected inputs while effectively preventing colorization. As shown in Fig.~\ref{fig:effect_imper}, NA introduces visible artifacts—especially on smooth regions like manga backgrounds and faces—whereas PAChroma maintains a natural appearance with minimal distortion.
%This improvement is attributed to the continuous Laplacian mask, which concentrates perturbations on high-frequency edge regions and avoids smooth areas where the human eye is most sensitive. 
Quantitatively, Tab.~\ref{tab:attack_eval} shows improved perceptual similarity for PAChroma, with PSNR increasing from 28.28 to 32.48 and SSIM from 0.72 to 0.94 on BigColor. 
Despite a modest drop in CF reduction (from 97.40\% to 82.80\%), colorization is still strongly suppressed.
These trends are consistent across models, indicating that the Laplacian mask enhances imperceptibility without sacrificing defense performance—crucial for manga, where large smooth areas like speech bubbles are common.

\textbf{Transferability.}
PAChroma achieves higher transferability in the black-box setting compared to NA with Laplacian masking (NA-Mask).
As shown in Fig.~\ref{fig:trans} and Tab.~\ref{tab:attack_eval},
% In the black-box setting, PAChroma enhances transferability by incorporating input transformation strategy.
NA-Mask yields only modest CF reduction (2.40\%–20.30\%), while PAChroma improves this to 13.60\%–28.30\%.
Although the CF reduction is modest compared to the white-box setting, PAChroma still produces perceptibly low-quality colorizations (Fig.~\ref{fig:trans}), effectively hindering malicious users from creating high-quality media for resale and other misuse.
These results suggest that PAChroma generalizes better across models by incorporating an input transformation strategy compared to NA-Mask, offering a more practical defense.

\textbf{Robustness.}  
PAChroma consistently outperforms NA-Mask in terms of robustness to post-processing.
As seen in Fig.~\ref{fig:robust} and Tab.~\ref{tab:attack_eval}, NA-Mask is highly vulnerable to common transformations like JPEG compression and random resized cropping (RRC), often resulting in partial colorization recovery.
In contrast, PAChroma retains significantly lower CF scores after post-processing—e.g., under JPEG 50\%, DDColor’s CF is 14.72 for PAChroma versus 24.41 for NA-Mask.
Although NA occasionally yields stronger suppression, PAChroma provides a better balance of imperceptibility and robustness, offering more dependable protection across models.

%As shown in Fig.\ref{tab:attack_eval}, the simplest baseline, NA, is vulnerable to compressions and random resized cropping. 
%Even if the effectiveness of NA is high, when JPEG compressions of quality 50\% are done the CF score increased to 12.7, 
%Whereas our proposed method keeps the suppression even under these transformations.

\begin{table}[t]
\centering
\caption{Comparison of image colorization prevention methods}
\renewcommand{\arraystretch}{1.4}
\resizebox{\columnwidth}{!}{  
\begin{tabular}{|l|c|c|c|c|}
\hline
\textbf{Method} & 
\textbf{\textcolor{mygreen}{Effectiveness}} & 
\textbf{\textcolor{myblue}{Imperceptibility}} & 
\textbf{\textcolor{mypurple}{Transferability}} & 
\textbf{\textcolor{myred}{Robustness}} \\\hline
%\textbf{Random} & \textbf{--} & \textbf{--} & \textbf{--} & \textbf{--} \\
Random Noise & 	\textbf{--} & \textbf{--} & \textbf{--} & \textbf{--} \\
NA & 	\textasciitilde90\% & \textbf{--} & \textbf{--} & \textbf{--} \\
NA-Mask~(ours) & \textasciitilde90\% & \checkmark  & \textbf{--}  & \textbf{--} \\
\textbf{PAChroma~(ours)} & 	\textasciitilde80\% & \checkmark & \checkmark & \checkmark \\
\hline
\end{tabular}
}
\label{tab:summary}
\end{table}

\textbf{Manga Domain.}
Tab.~\ref{tab:manga_eval} demonstrates that PAChroma achieves strong effectiveness, imperceptibility, and robustness against manga colorization models. 
However, its transferability across models remains limited, likely due to the unique characteristics of manga—namely, large flat regions with minimal texture, which constrain perturbation flexibility and hinder generalization.
Higher image resolution compared to natural image datasets may further amplify this challenge.

Moreover, our method currently targets fully automatic colorization. Extending to user-guided approaches (e.g., scribble- or text-based) as well as to higher-resolution and more computationally efficient settings remains an open challenge.

%\section{Limitations}
%Our method focuses on fully automatic colorization. Extending the adversarial attack framework to user-guided or semi-automatic models (e.g., scribble-based or text-driven colorization) remains an open challenge for future work.

\section{Conclusions}
We introduce \textit{Uncolorable Examples}—grayscale images with imperceptible perturbation that resist AI colorization.
Generated by our method \textit{PAChroma}, which combines input transformations and a Laplacian mask, they suppress colorization while preserving appearance.
PAChroma is effective and imperceptible in white-box settings, with moderate transferability and robustness (Tab.~\ref{tab:summary}), laying the foundation for protecting content from unauthorized generative manipulation.

\section{Acknowledgments}
This work was partially supported by JSPS KAKENHI Grants JP21H04907 and JP24H00732, by JST CREST Grant JPMJCR20D3 including AIP challenge program, by JST AIP Acceleration Grant JPMJCR24U3, and by JST K Program Grant JPMJKP24C2 Japan.

%\printbibliography
\bibliographystyle{IEEEtran}
\bibliography{mybib}

\begin{thebibliography}{10}
\providecommand{\url}[1]{#1}
\csname url@samestyle\endcsname
\providecommand{\newblock}{\relax}
\providecommand{\bibinfo}[2]{#2}
\providecommand{\BIBentrySTDinterwordspacing}{\spaceskip=0pt\relax}
\providecommand{\BIBentryALTinterwordstretchfactor}{4}
\providecommand{\BIBentryALTinterwordspacing}{\spaceskip=\fontdimen2\font plus
\BIBentryALTinterwordstretchfactor\fontdimen3\font minus \fontdimen4\font\relax}
\providecommand{\BIBforeignlanguage}[2]{{%
\expandafter\ifx\csname l@#1\endcsname\relax
\typeout{** WARNING: IEEEtran.bst: No hyphenation pattern has been}%
\typeout{** loaded for the language `#1'. Using the pattern for}%
\typeout{** the default language instead.}%
\else
\language=\csname l@#1\endcsname
\fi
#2}}
\providecommand{\BIBdecl}{\relax}
\BIBdecl

\bibitem{huang2022unicolorunifiedframeworkmultimodal}
Z.~Huang, N.~Zhao, and J.~Liao, ``Unicolor: A unified framework for multi-modal colorization with transformer,'' 2022.

\bibitem{wu2022vividdiverseimagecolorization}
Y.~Wu, X.~Wang, Y.~Li, H.~Zhang, X.~Zhao, and Y.~Shan, ``Towards vivid and diverse image colorization with generative color prior,'' 2022.

\bibitem{mainichi2025godzilla}
T.~Mainichi. (2025, 6) Man arrested in japan for selling ai-colorized pirated 1954 'godzilla' film.

\bibitem{DeOldify2021}
J.~Antic, ``{DeOldify}: Deep learning for image colorization and restoration,'' 2021, gitHub, \url{https://github.com/jantic/DeOldify}.

\bibitem{dakini2024animecolor}
Dakini, AIEMMU, and A.~Regmi, ``Animecolordeoldify: Deoldify-based colorization for anime, sketch, and manga,'' 2024, gitHub, https://github.com/Dakini/AnimeColorDeOldify.

\bibitem{kang2023ddcolorphotorealisticimagecolorization}
X.~Kang, T.~Yang, W.~Ouyang, P.~Ren, L.~Li, and X.~Xie, ``Ddcolor: Towards photo-realistic image colorization via dual decoders,'' 2023, https://arxiv.org/abs/2212.11613.

\bibitem{mangacolorv2_2022}
{Manga Colorization V2}, ``Manga colorization v2,'' https://github.com/qweasdd/manga-colorization-v2, 2022.

\bibitem{kim2022bigcolorcolorizationusinggenerative}
G.~Kim, K.~Kang, S.~Kim, H.~Lee, S.~Kim, J.~Kim, S.-H. Baek, and S.~Cho, ``Bigcolor: Colorization using a generative color prior for natural images,'' 2022.

\bibitem{szegedy2013intriguing}
C.~Szegedy, W.~Zaremba, I.~Sutskever, J.~Bruna, D.~Erhan, I.~Goodfellow, and R.~Fergus, ``Intriguing properties of neural networks,'' \emph{arXiv preprint arXiv:1312.6199}, 2013.

\bibitem{madry2018pgd}
A.~Madry, A.~Makelov, L.~Schmidt, D.~Tsipras, and A.~Vladu, ``Towards deep learning models resistant to adversarial attacks,'' in \emph{International Conference on Learning Representations (ICLR)}, 2018.

\bibitem{dong2018boostingadversarialattacksmomentum}
Y.~Dong, F.~Liao, T.~Pang, H.~Su, J.~Zhu, X.~Hu, and J.~Li, ``Boosting adversarial attacks with momentum,'' 2018.

\bibitem{goodfellow2015explainingharnessingadversarialexamples}
I.~J. Goodfellow, J.~Shlens, and C.~Szegedy, ``Explaining and harnessing adversarial examples,'' 2015.

\bibitem{9096939}
C.-Y. Yeh, H.-W. Chen, S.-L. Tsai, and S.-D. Wang, ``Disrupting image-translation-based deepfake algorithms with adversarial attacks,'' in \emph{2020 IEEE Winter Applications of Computer Vision Workshops (WACVW)}, 2020, pp. 53--62.

\bibitem{huang2021cmuawatermarkcrossmodeluniversaladversarial}
H.~Huang, Y.~Wang, Z.~Chen, Y.~Zhang, Y.~Li, Z.~Tang, W.~Chu, J.~Chen, W.~Lin, and K.-K. Ma, ``Cmua-watermark: A cross-model universal adversarial watermark for combating deepfakes,'' 2021, https://arxiv.org/abs/2105.10872.

\bibitem{wang2022antiforgerystealthyrobustdeepfake}
R.~Wang, Z.~Huang, Z.~Chen, L.~Liu, J.~Chen, and L.~Wang, ``Anti-forgery: Towards a stealthy and robust deepfake disruption attack via adversarial perceptual-aware perturbations,'' 2022, https://arxiv.org/abs/2206.00477.

\bibitem{xie2018mitigatingadversarialeffectsrandomization}
C.~Xie, J.~Wang, Z.~Zhang, Z.~Ren, and A.~Yuille, ``Mitigating adversarial effects through randomization,'' 2018, https://arxiv.org/abs/1711.01991.

\bibitem{guo2018counteringadversarialimagesusing}
C.~Guo, M.~Rana, M.~Cisse, and L.~van~der Maaten, ``Countering adversarial images using input transformations,'' 2018, https://arxiv.org/abs/1711.00117.

\bibitem{colorfulness}
D.~Hasler and S.~Suesstrunk, ``Measuring colourfulness in natural images,'' \emph{Proceedings of SPIE - The International Society for Optical Engineering}, vol. 5007, pp. 87--95, 06 2003.

\bibitem{shen2021structure}
C.~Shen, Y.~Dong, H.~Su, and J.~Zhu, ``Structure-preserving transformation for adversarial example generation,'' in \emph{Proceedings of the IEEE/CVF International Conference on Computer Vision (ICCV)}, 2021, pp. 6890--6900.

\bibitem{watson1997contrast}
A.~B. Watson and J.~A. Solomon, ``Model of visual contrast gain control and pattern masking,'' \emph{JOSA A}, vol.~14, no.~9, pp. 2379--2391, 1997.

\bibitem{deng2009imagenet}
J.~Deng, W.~Dong, R.~Socher, L.-J. Li, K.~Li, and L.~Fei-Fei, ``Imagenet: A large-scale hierarchical image database,'' in \emph{IEEE Conference on Computer Vision and Pattern Recognition (CVPR)}, 2009, pp. 248--255.

\bibitem{xie-2018-manga}
M.~Xie, C.~Li, X.~Liu, and T.-T. Wong, ``Manga filling style conversion with screentone variational autoencoder,'' \emph{ACM Transactions on Graphics (SIGGRAPH Asia 2020 issue)}, vol.~39, no.~6, pp. 226:1--226:15, 12 2020.

\end{thebibliography}

\newpage

\appendix
\section{Supplementary Material}
We include additional results and insights in the appendix.

\subsection{Additional Visual Results}
\begin{figure*}
\centering
\includegraphics[width=\linewidth]{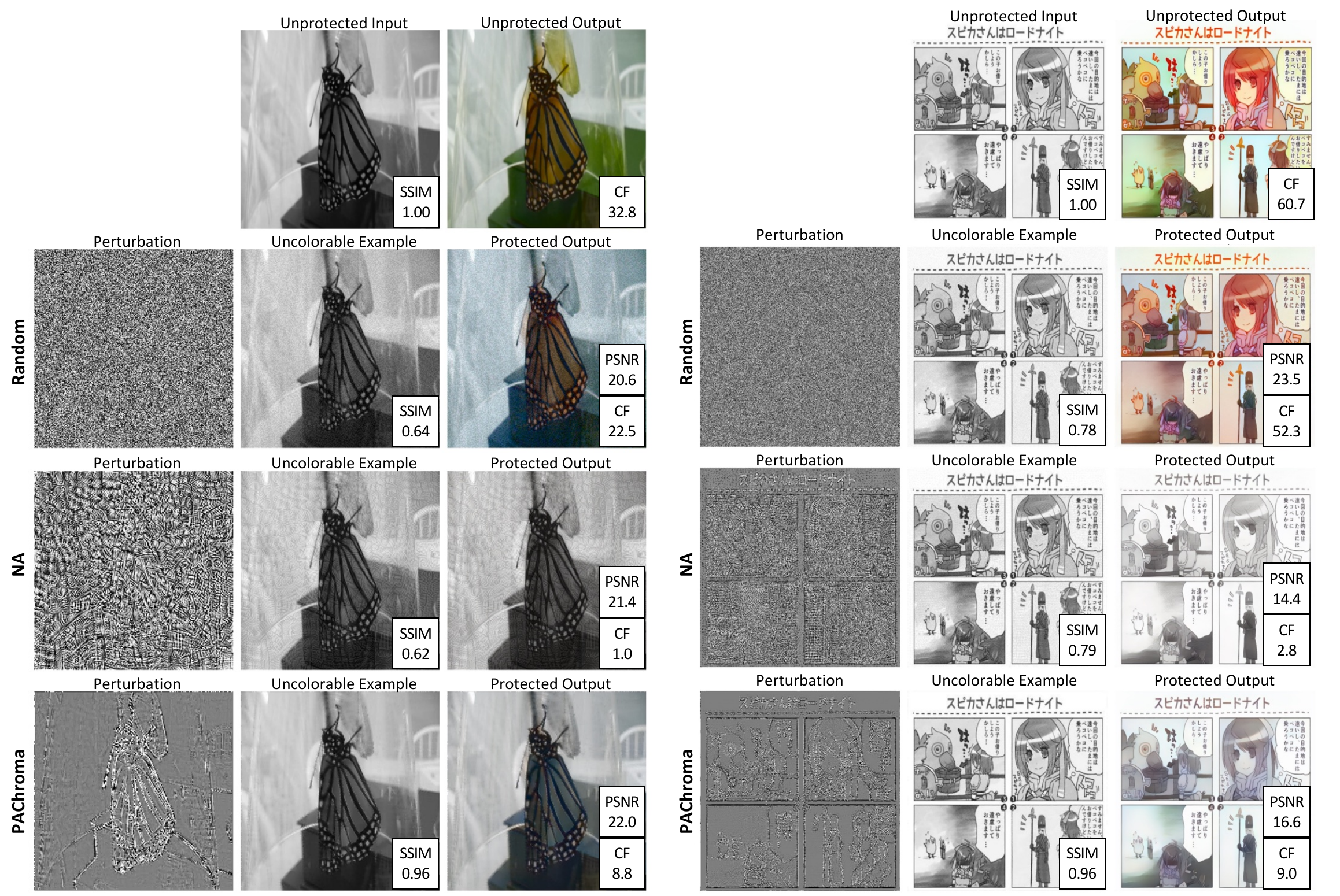}
\caption{Additional visual results comparing NA, NA-Mask and PAChroma for imperceptibility.}
\label{fig:appendix_imp}
\end{figure*}

\begin{figure*}
\centering
\includegraphics[width=\linewidth]{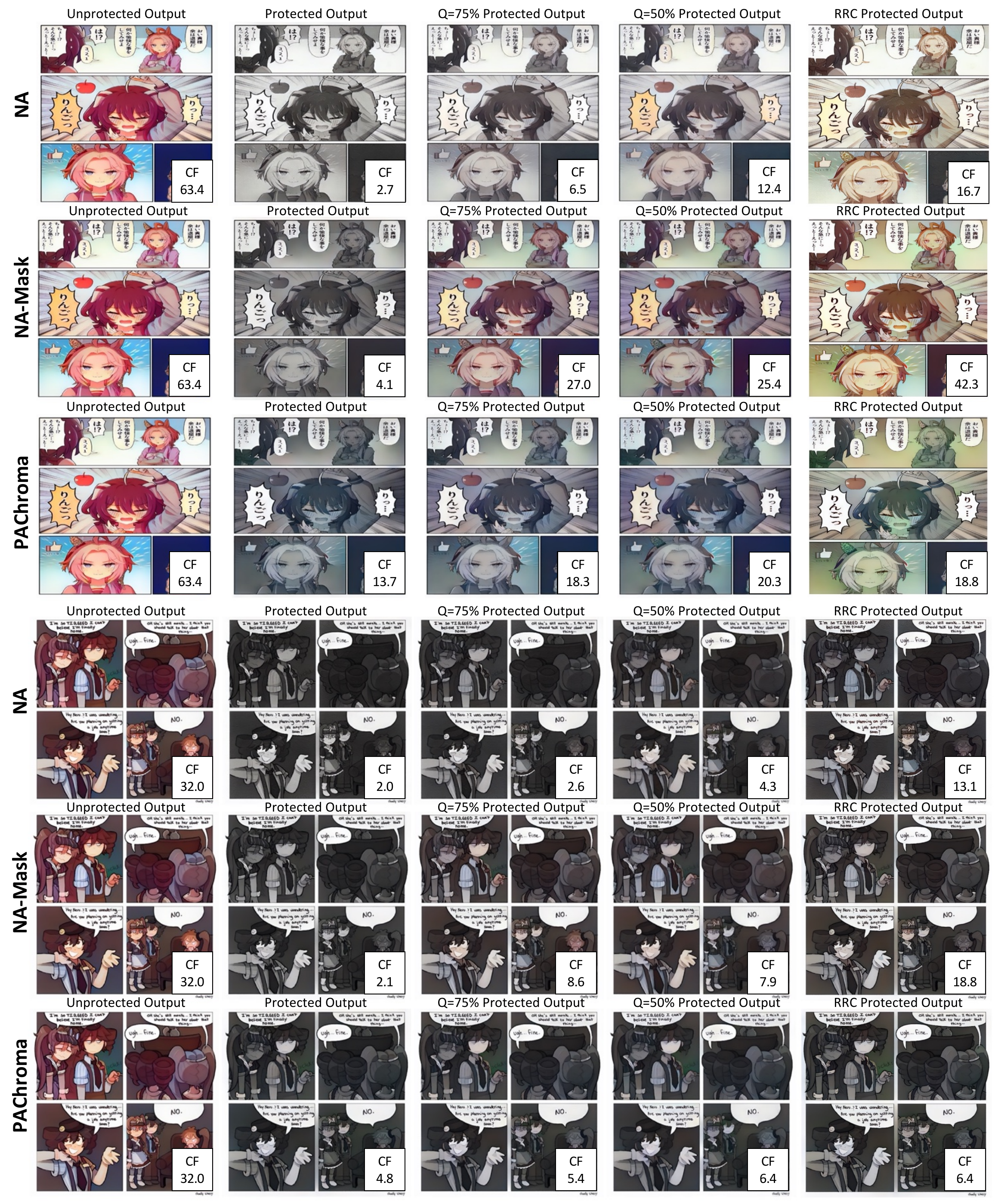}
\caption{Additional visual results comparing NA, NA-Mask and PAChroma for robustness.}
\label{fig:appendix_rob}
\end{figure*}

Fig.~\ref{fig:appendix_imp} illustrates how PAChroma effectively prevents colorization while remaining imperceptible to human viewers.
Fig.\ref{fig:appendix_rob} demonstrates its robustness in piracy contexts, particularly under JPEG compression and random resizing.

\subsection{Further Insights}
We observed that Laplacian edge-weighted masking improves imperceptibility by
focusing perturbations on high-frequency regions, while block-wise
transformations promote robustness against augmentation. These insights were
noted consistently across both natural and manga images.

Experiments under both $L_2$ and $L_\infty$ bounds show that enlarging the perturbation budget $\epsilon$ improves transferability and effectiveness, but reduces imperceptibility. This highlights a clear trade-off between effectiveness and imperceptibility. Considering this trade-off, we choose parameters that balance all criteria.
\end{document}